\documentclass[conference]{IEEEtran}
\IEEEoverridecommandlockouts
\usepackage{cite}
\usepackage{amsmath,amssymb,amsfonts}
\usepackage{algorithmic}
\usepackage{graphicx}
\usepackage{textcomp}
\usepackage{balance}
\usepackage[table, dvipsnames]{xcolor}
\usepackage[utf8]{inputenc} 
\usepackage[T1]{fontenc}    
\usepackage{hyperref}       
\usepackage{url}            
\usepackage{booktabs}       
\usepackage{nicefrac}       
\usepackage{microtype}      
\usepackage{tabularx}
\usepackage{multirow}
\usepackage{subfigure}
\usepackage{graphicx}
\usepackage{pgfplots}
\pgfplotsset{compat=1.16}

\usepackage{tcolorbox}
\definecolor{lightgray}{gray}{0.9}

\def\BibTeX{{\rm B\kern-.05em{\sc i\kern-.025em b}\kern-.08em
    T\kern-.1667em\lower.7ex\hbox{E}\kern-.125emX}}
\begin{document}

\DeclareRobustCommand*{\IEEEauthorrefmark}[1]{%
  \raisebox{0pt}[0pt][0pt]{\textsuperscript{\footnotesize #1}}%
}

\title{One-Shot Federated Learning with Classifier-Free Diffusion Models\\}
\author{\IEEEauthorblockN{Obaidullah Zaland\IEEEauthorrefmark{1}\textsuperscript{*},
Shutong Jin\IEEEauthorrefmark{2}\textsuperscript{*},
Florian T. Pokorny\IEEEauthorrefmark{2} and  
Monowar Bhuyan\IEEEauthorrefmark{1}}

\IEEEauthorblockA{\IEEEauthorrefmark{1}Department of Computing Science,
        Ume\r{a} Unviersity, Ume\r{a}, SE-90187, Sweden\\
emails: \{ozaland, monowar\}@cs.umu.se}

\IEEEauthorblockA{\IEEEauthorrefmark{2}KTH Royal Institute of Technology, Stockholm, SE-10044, Sweden\\
emails: \{shutong, fpokorny\}@kth.se}

\thanks{This work was partially supported by the Wallenberg AI, Autonomous Systems and Software Program (WASP) funded by the Knut and Alice Wallenberg Foundation via the WASP NEST project “Intelligent Cloud Robotics for Real-Time Manipulation at Scale.” The computations and data handling essential to our research were enabled by the supercomputing resource Berzelius provided by the National Supercomputer Centre at Linköping University and the gracious support of the Knut and Alice Wallenberg Foundation.}
}


\maketitle
\begin{abstract}

Federated learning (FL) enables collaborative learning without data centralization but introduces significant communication costs due to multiple communication rounds between clients and the server. One-shot federated learning (OSFL) addresses this by forming a global model with a single communication round, often relying on the server's model distillation or auxiliary dataset generation - mostly through pre-trained diffusion models (DMs). Existing DM-assisted OSFL methods, however, typically employ classifier-guided DMs, which require training auxiliary classifier models at each client, introducing additional computation overhead. This work introduces OSCAR (One-Shot Federated Learning with Classifier-Free Diffusion Models), a novel OSFL approach that eliminates the need for auxiliary models. OSCAR uses foundation models to devise category-specific data representations at each client which are integrated into a classifier-free diffusion model pipeline for server-side data generation. In our experiments, OSCAR outperforms the state-of-the-art on four benchmark datasets while reducing the communication load by at least $\mathbf{99}\%$\footnote{\url{https://github.com/obaidullahzaland/oscar}}.

\end{abstract}


\renewcommand{\thefootnote}{\fnsymbol{footnote}}
\footnotetext[1]{Equal contributions.}
\section{Introduction}

Federated Learning (FL) ~\cite{mcmahan2017communication} is a decentralized machine learning (ML) training methodology that enables multiple clients to collaboratively train a global model without moving the data to a central location, thus addressing concerns about data privacy and ownership in the age of growing data privacy regulations. This has led to the application of FL to various domains, including autonomous vehicles ~\cite{xue2024spatial}, the Internet of Things (IoT) ~\cite{wang2022blockchain}, and healthcare ~\cite{banerjee2020multi}. However, since the participating clients usually own data that is not independent and identically distributed (non-IID) data, the task of training an \textit{optimal} global model typically requires multiple communication rounds between the clients and the server, causing high communication overhead ~\cite{kairouz2021advances}. Several strategies, including client selection ~\cite{singhal2024greedy}, update compression ~\cite{lan2023improved}, and update dropping ~\cite{zhou2024accelerating}, have been proposed to reduce the communication load in each communication round. However, these strategies still require client synchronization and suffer under non-IID data distribution across clients, as local client models can drift from the global model at each communication round ~\cite{shi2022optimization}. 
 
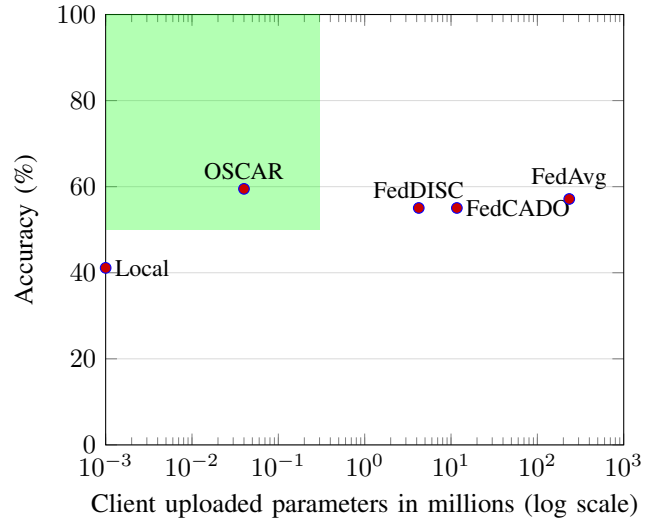
\begin{figure}
    \centering
    \begin{tikzpicture}
  \begin{axis}[
    xmode=log, xmin=0.001, xmax=1000,
    xlabel={Client uploaded parameters in millions (log scale)},
    ylabel={Accuracy (\%)},
    ymin=0, ymax=100,
    xmajorgrids=false,
    xminorgrids=false,
    ymajorgrids=true,
    yminorgrids=false,
    grid style={line width=.1pt, draw=gray!30},
    title={}
  ]
  
  \addplot [
    draw=none,
    fill=green,
    fill opacity=0.3
  ]
  coordinates {
    (0.001,50)
    (0.3,50)
    (0.3,100)
    (0.001,100)
    (0.001,50)
  };

  \addplot+[
    only marks,
    mark=*,
    color=blue
  ]
  coordinates {
    (0.001,41.13)
    (234,57.14)
    (11.69,55.05)
    (4.23,55.05)
    (0.04,59.49)
  };

  \node[anchor=west, font=\small, color=black] at (axis cs:0.001,41.13) {Local};
  \node[anchor=south, font=\small, color=black] at (axis cs:234,57.14) {FedAvg};
  \node[anchor=west, font=\small, color=black] at (axis cs:11.69,55.05) {FedCADO};
  \node[anchor=south, font=\small, color=black] at (axis cs:4.23,55.05) {FedDISC};
  \node[anchor=south, font=\small, color=black] at (axis cs:0.04,59.49) {OSCAR};

  \end{axis}
\end{tikzpicture}
    \caption{Uploaded parameters by each client and accuracy for various algorithms on the OpenImage dataset and ResNet-18.}
    \label{fig:communicaiton}
\end{figure}

One-shot federated learning (OSFL)~\cite{guha2019one} offers an alternative, where the global model is learned through a single communication round between the clients and the server. OSFL can also reduce the impact of heterogeneous data distribution, as the global model is not directly formed from the local models. The single communication round can further benefit in scenarios where FL suffers from client dropout or stragglers (i.e., clients that communicate slowly)~\cite{yang2023one}. Existing OSFL approaches~\cite{zhang2022dense, zhou2020distilled, yang2024feddeo} rely on auxiliary dataset generation or knowledge distillation to form the global model. Knowledge distillation approaches usually require an auxiliary public dataset as the knowledge transfer medium~\cite{zhou2020distilled}. On the other hand, dataset generation methods employ pre-trained generative models (e.g., diffusion models) to generate \textit{new} data for training the global model. DMs are pre-trained with vast amounts of data, and \textit{with proper guidance}, these DMs can generate realistic images that resemble a desired distribution. DMs can have an immense impact on FL, as new data resembling the clients' distribution can be generated without access to the raw dataset of the participating clients.

Current studies incorporating diffusion models in OSFL~\cite{yang2023one, mendieta2024navigating, yang2024feddeo, zhang2023federated} utilize classifier-guided DMs. Employing classifier-guided DMs requires auxiliary classifier training at each client, introducing computational and communication burdens. Furthermore, in some cases, the diffusion model needs to be downloaded to the clients~\cite{yang2024feddeo}. Classifier-free DMs~\cite{ho2022classifier} solve these challenges by integrating the conditioning directly into the model, which is also adopted by most of the prevalent image generative models~\cite{betker2023improving, rombach2022high}. Foundation models (FMs)~\cite{li2023blip, achiam2023gpt} can be employed for the encoding generation that can be used as conditioning without training or fine-tuning. Replacing the classifier models with encodings significantly reduces the client upload size compared to classifier-guided DM-based OSFL approaches, as shown in \textit{Fig} \ref{fig:communicaiton}. The seamless integration of pre-trained FMs and DMs in OSFL simplifies the overall framework, reduces communication load, and enhances scalability and efficiency across heterogeneous client datasets.

In this work, we present \textbf{OSCAR}, \textbf{O}ne-\textbf{S}hot federated learning with \textbf{C}l\textbf{A}ssifier-F\textbf{R}ee diffusion models. OSCAR leverages the strengths of FMs and a classifier-free diffusion model to train a global FL model in a single communication round between the clients and the server. OSCAR relies on each participating party's category-specific encodings to generate data through classifier-free DMs. The generated data is then used to train the global model on the server. By removing the need for classifier training at each client, OSCAR reduces the client upload size by 99\% compared to current state-of-the-art (SOTA) DM-assisted OSFL approaches. In addition to reducing the communication overhead at each client, OSCAR outperforms existing SOTA on four different benchmarking datasets.

\section{Prior Work}

\subsection{One-Shot Federated Learning}
Existing OSFL approaches can be divided into two categories based on their methodology. The first category utilizes knowledge distillation to learn a global model through either data distillation~\cite{zhou2020distilled} or model distillation~\cite{li2020practical}. In distilled one-shot federated learning (DOSFL~\cite{zhou2020distilled}, the clients share distilled synthetic data with the server, which is utilized for global model training. FedKT~\cite{li2020practical} utilizes a public auxiliary dataset and student-teacher models trained by clients to learn a global student model. The second category of methods uses auxiliary data generation at the server based on intermediary information shared by the clients. DENSE ~\cite{zhang2022dense} trains a generator model on local classifiers, later used to generate auxiliary data for global model training. In FedCVAE~\cite{heinbaugh2023data}, the server aggregates the decoder part of the conditional variational encoders (CVAE) trained at each client and generates auxiliary data for the global model. FedDiff~\cite{mendieta2024navigating} aggregates locally trained diffusion models to form a global diffusion model for data generation. FedCADO~\cite{yang2023one} utilizes classifiers trained at each client to generate data for global model training via classifier-guided pre-trained diffusion models (DMs). FedDISC~\cite{yang2024exploring} utilizes data features for data generation via pre-trained DMs. 

\subsection{Federated Learning with Foundation Models}

The emergence of foundation models (FMs), both large language models (LLMs)~\cite{achiam2023gpt} and vision language models (VLMs)~\cite{li2023blip}, has impacted the landscape of machine learning. The application of these FMs, however, has not yet been fully explored in FL. Yu et al., ~\cite{yu2024federated}, and Charles et al., ~\cite{charles2024towards} explore training FMs in FL setups. PromptFL~\cite{guo2023promptfl} investigates prompt learning for FMs under data scarcity in FL settings. FedDAT ~\cite{chen2024feddat} proposes a federated fine-tuning approach for multi-modal FMs. FedPCL~\cite{tan2022federated} integrates FMs into the traditional FL process to act as class-wise prototype extractors. While FMs have the potential to mitigate data heterogeneity and communication load in FL, their full potential has not been utilized in FL settings. 

\section{Preliminaries}



\subsection{Diffusion Models}
Denoising Diffusion Probabilistic Models (DDPMs)~\cite{ho2020denoising} employ a U-Net architecture~\cite{ronneberger2015u}, denoted as $\epsilon_{\theta}$, to model data distribution $x \sim q(x_0)$. For any given timestamp $t \in \{0, \ldots, T\}$
, during the \textit{forward process}, Gaussian noise $\mathbf{I}$ is progressively added according to:
\begin{equation}
q(x_t|x_{t-1}) = \mathcal{N}(x_t; \sqrt{1-\beta_t}x_{t-1},\; \beta_t \mathbf{I}),
\end{equation}

with $\beta_t$ as a learned variance scheduler. In the \textit{reverse process}, $x_0$ is sampled from:

\begin{equation}
p_{\theta}(x_{t-1} | x_t) \sim \mathcal{N}(x_{t-1}; \mu_{\theta}(x_t, t),\; \Sigma_{\theta}(x_t,\;t)),
\end{equation}

where $\mu_{\theta}(x_t, t)$ is derived from $\epsilon_\theta (x_t, t)$, and $\Sigma_{\theta}(x_t,t)$ is a time-dependent constant. 

When a conditioning signal $y$, such as text, is added, the network is trained to minimize:
\begin{equation}
\begin{split}
    \mathcal{L}_{t}(\theta) = \mathbb{E}_{z_0\sim q(\mathcal{E}(x_{0})), \;\epsilon\sim\mathcal{N}(0,\;\mathbf{I}),\;t}\left[ \left\|\epsilon - \hat{\epsilon}_t\right\|^{2}\right], \\
    \hat{\epsilon}_t = \epsilon_\theta (x_t,\; t,\; y),
\end{split}
\end{equation}
where $\hat{\epsilon}_t$ provides an estimate of the score function used for data generation during the reverse process, and \( \mathcal{E}(x_0) \) denotes the encoded representation of the original input \( x_0 \).

\textbf{Classifier-Guided Models}~\cite{dhariwal2021diffusion} generate conditional samples by combining the diffusion model's score estimate with the input gradient of a classifier's log probability:
\begin{equation}
    \hat{\epsilon}_t = \epsilon_\theta(x_t, t, y) - s \sigma_t \nabla_{x_t} \log p(y|x_t),
\end{equation}
where \( p(y \mid x_t)\) is the probability of class \(y\) with respect to the input \(x_t\), obtained from the classifier. \(\sigma_t\) denotes the noise scale at timestep \(t\), while $s$ controls the influence of the classifier's guidance on the classifier's output.

\textbf{Classifier-Free Models}~\cite{ho2022classifier}, in contrast, combine the score estimate from a conditional diffusion model with that of a jointly trained unconditional model, guiding the generation by their difference:
\begin{equation}
    \hat{\epsilon}_t = (1 + s)\epsilon_\theta(x_t, t, y) - s\epsilon_\theta(x_t, t, \emptyset),
\end{equation}
where \(\epsilon_\theta(x_t, t, \emptyset)\) represents prediction without conditioning.

In conclusion, classifier-guided models rely on a trained classifier to guide the generative process by predicting the likelihood that the generated output matches a given text description. Classifier-free Models integrate text conditioning directly into the generative process, eliminating the need for an external classifier. This approach has driven recent trends in training text-driven generative models, such as DALL·E~\cite{ramesh2022hierarchical} and Stable Diffusion~\cite{rombach2022high}.
\section{OSCAR: One Shot Federated Learning with Classifier-Free Diffusion Models}
\begin{figure*}[!t]
    \centering
    \includegraphics[width=0.83\textwidth]{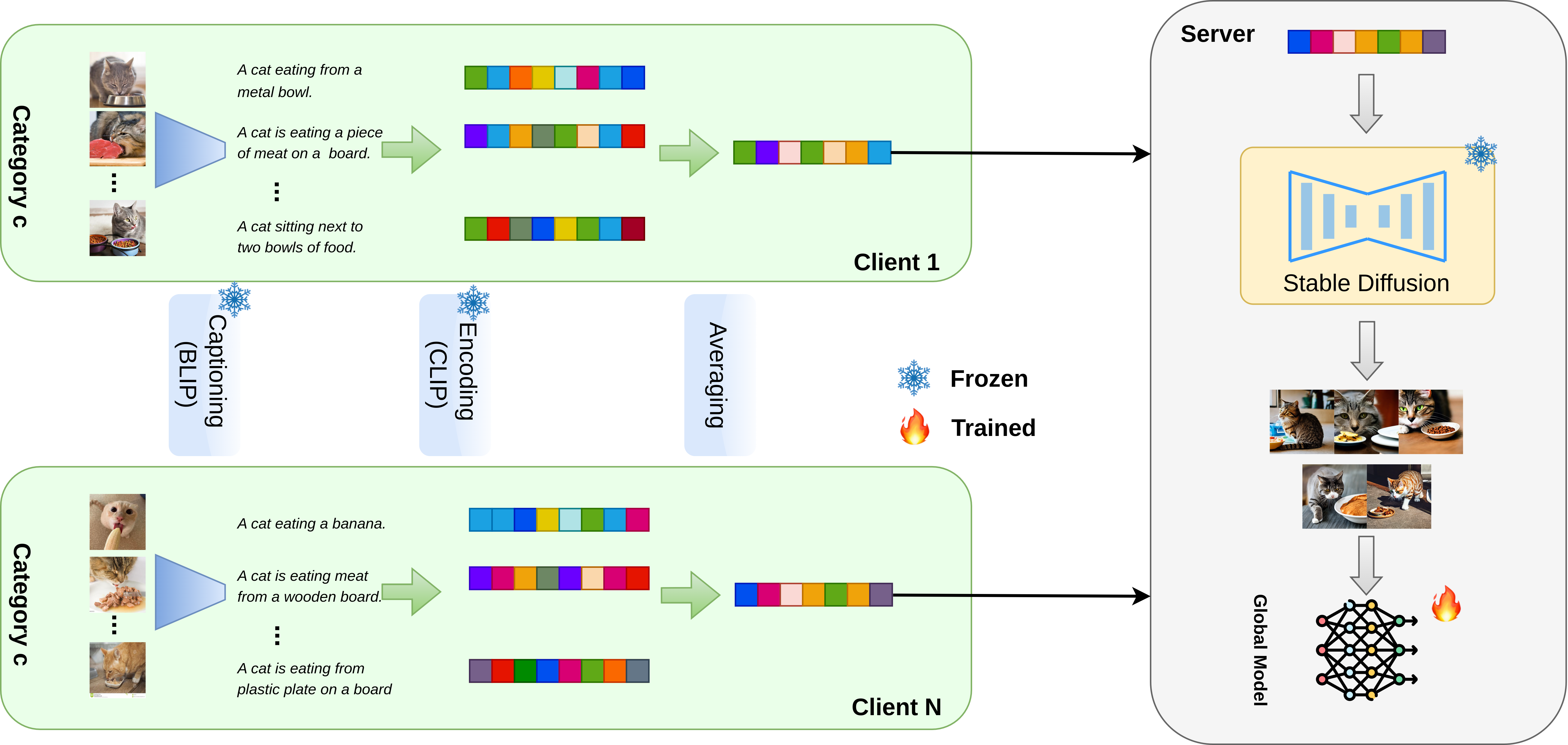}
    \caption{An illustration of the proposed OSCAR pipeline, where BLIP~\cite{li2022blip}, CLIP~\cite{radford2021learning} Text Encoder, and Stable Diffusion~\cite{rombach2022high} are all used with frozen weights and in a zero-shot manner.}
    \label{structure}
\end{figure*}
The traditional FL setup consists of a central server and a set of clients $\mathcal{R}$, where each client $r\in\mathcal{R}$ trains a local model $\mathbf{w}_r$ on its local dataset $\mathcal{D}_r$ in each iteration and communicates it to the server. The server is responsible for forming a global model $\mathbf{w}$ from the local client models through an aggregation function like federated averaging (FedAvg). Considering the intrinsic data heterogeneity in FL, traditional FL algorithms require multiple communication rounds between the clients and the server to form a global model, leading to significant communication overhead. 

To reduce the communication load in the FL setup, we propose OSCAR, a novel one-shot federate learning (OSFL) approach. OSCAR integrates foundation and pre-trained classifier-free generative models, specifically Stable Diffusion~\cite{rombach2022high}, and learns a global model from the clients' category-specific representations. OSCAR facilitates global model learning within a single communication round under a non-IID data distribution among the clients. As illustrated in \textit{Fig.}~\ref{structure}, the OSCAR pipeline can be divided into four steps: \textbf{(1)} generating descriptions of the client's client-specific data, \textbf{(2)} encoding features from the generated descriptions, and transmitting them to the server \textbf{(3)} transmitting the category-specific representations to the server, and \textbf{(4)} generating data on the server to facilitate final model training. 
\paragraph{Description Generation and Text Encoding} Unlike existing DM-assisted OSFL approaches, OSCAR eliminates classifier training and utilizes the clients' category-specific data features as conditioning for the DM. Each client follows a two-step approach to generate category-specific encodings. First, the client uses a VLM, specifically BLIP~\cite{li2022blip}, to generate textual descriptions for all their images. Then, the client uses a text encoder, specifically CLIP~\cite{radford2021learning}, denoted as $\text{CLIP}_\text{Text}$, to generate category-specific text encodings from the textual descriptions, as shown in \textit{Eq.}~\ref{blipclip}. A classifier-free diffusion model can utilize the CLIP encodings directly as text conditioning to generate new data.

\begin{equation}
\begin{split}
    y_{cn} &= \textsc{CLIP}_{\text{Text}} \left( \textsc{BLIP}(x_{cn}) \right), \\
    &\quad \text{for } c = 1, \dots, C \text{ and } n = 1, \dots, N
\end{split}
\label{blipclip}
\end{equation}
where $C$ represents the number of categories for the client, $N$ is the number of category-specific images at the party, and $y_{cn}$ denotes the encoded text corresponding to the input $x_{cn}$.

\paragraph{Client Representation and Server Data Synthesis} 
Each client averages the category-specific encodings to form a unified representation for the specific category. Despite its simplicity, averaging the category-specific text encodings aligns well with the classifier-free approach:

\begin{equation}
    \bar{y}_c = \frac{1}{N} \sum_{n=1}^{N} y_{cn}
\end{equation}
The averaged feature \(\bar{y}_c\) for each category $c$ is then sent directly to the server to initiate classifier-free sampling:
\begin{equation}
    \hat{\epsilon}_T(\bar{y}_c) = (1 + s) \epsilon_\theta(x_T, T, \bar{y}_c) - s \epsilon_\theta(x_T, T, \emptyset)
\end{equation}
where the guidance scale \(s\) is fixed at 7.5. In this process, \(x_T\) (with \(T\) set to 50) is randomly sampled from the noise space, and the subsequent sampling is carried out according to:
\begin{equation}
    x_{T-1} = \frac{1}{\sqrt{\alpha_T}} \left( \sqrt{\alpha_T} x_T - \hat{\epsilon}_T(\bar{y}_c) \right) + \sigma_T \mathcal{N}(0, \mathbf{I}) 
\end{equation}
Starting from random noise $x_T$, the model iteratively refines the image through the sequence $ x_T \rightarrow x_{T-1} \rightarrow \dots \rightarrow x_0$, guided by the category-specific encoding $\bar{y}_c$, to produce the final output image $x_0$ within the client's distribution. 
In our comparisons to SOTA models, replacing the classifier model with the clients’ category-specific representations has the advantage of reducing each client's upload size by at least 99
The server generates \textbf{ten} images for each category-specific client representation to form a global synthesized dataset $\mathcal{D}_{syn}$, with $10 \times |\mathcal{R}| \times C$ new images, where $|\mathcal{R}|$ is the number of clients and $C$ is the number of categories. As the global dataset $\mathcal{D}_{syn}$ is constructed based on the category-specific representations from individual clients, it effectively captures the heterogeneous data distribution present at each client.

\paragraph{Model Training} After generating the global synthesized dataset $\mathcal{D}_{syn}$, the server trains a centralized model $\mathbf{w}$, specifically a ResNet-18 classifier, on the synthesized dataset. This approach not only reduces the dependency on client synchronization and availability at each training and communication round but also ensures that the model can generalize well to non-IID data across clients. The server communicates the global model $\mathbf{w}$ to all clients after training, to be later used for inference locally.

\section{Experimental Setup}
\paragraph{Datasets}
\begin{itemize}
    \item \textbf{NICO++} ~\cite{zhang2023nico++}: NICO++ contains images of size $224 \times 224$ from $60$ different categories, across six domains. The dataset has two settings. In \textit{Common NICO++}, all the categories share the same six domains: [autumn, dim, grass, outdoor, rock, and water]. On the other hand, the \textit{Unique NICO++} contains different domains for each category. 

    \item \textbf{DomainNet} ~\cite{peng2019moment}: The DomainNet dataset contains images of 345 categories over six domains. This work uses a subset of the DomainNet dataset with 90 categories. 

    \item \textbf{OpenImage} ~\cite{kuznetsova2020open}: OpenImage is a multi-task dataset with over 1.7 million images across 600 categories. This work uses a subset of 120 categories from the dataset following the pre-processing in ~\cite{yang2023one}.
 \end{itemize}

\paragraph{Data Division} The data has been divided among all parties in a non-IID manner. Specifically, the data is \textbf{feature distribution skewed}, as each client owns data about a single domain from each category in the NICO++ and DomainNet datasets. For Openimage, the classes have been divided into six similar subgroups, where each client owns a single category from each subgroup. The number of clients is fixed to \textbf{six}, aligning with the number of domains in all the datasets. 

\paragraph{Baselines} Local learning, federated learning, and state-of-the-art DM-assisted OSFL approaches have been considered as baselines for comparison against OSCAR. Each client trains a local standalone model on its data in local training. FedAvg~\cite{mcmahan2017communication}, FedProx~\cite{li2020federated}, and FedDyn~\cite{acar2021federated} consider traditional FL setups with minor variations in the local objective and aggregation functions. FedDISC~\cite{yang2024exploring} and FedCADO~\cite{yang2023one} are DM-assisted OSFL approaches and train auxiliary models for image generation at the server.

\section{Results and Analysis}

\paragraph{Main Results} In this section, experimental results are provided to compare OSCAR against baselines. The main comparisons are carried out on the four benchmarking datasets. In local and traditional FL settings, original images are used for training client models, while in FedCADO, FedDISC, and OSCAR, the global models are trained with synthetic data. The test set images, however,  are the actual dataset test images in all the experiments. We have compared OSCAR against two SOTA DM-assisted approaches, FedCADO~\cite{yang2023one} and FedDISC~\cite{yang2024exploring}, alongside traditional FL algorithms and local training. The experiments consider accuracy as the performance measure, calculated as the number of images classified correctly by the trained model divided by the number of images in the test set. Specifically, we only consider \textit{top-1} accuracy.  OSCAR performs better than the baselines on all the considered datasets, as shown in \textit{Table}. \ref{tab:results}. Aside from superior \textit{average} accuracy on the overall test set, OSCAR also performs better than the SOTA on domain-specific test sets. As we have assigned each domain to a single client, we consider the domain-specific test sets as client-specific test sets. All the experiments were carried out with the ResNet-18 classifier network, and the number of images per category for each client was set to 30. 

Like other DM-assisted OSFL approaches, OSCAR performs better on datasets that consist of real images (i.e., NICO++ and OpenImage). The difference is evident in the two domains (sketch and clipart) corresponding to client2 and client3 in the DomainNet dataset. While OSCAR performs lower than average in these domains, FL algorithms struggle similarly. 

\begin{table*}[!htb]
    \centering
    \caption{Accuracy (in \%) on the test set for the baselines and OSCAR on four benchmarking datasets. The best results are in bold.}
    \begin{tabularx}{\textwidth}{m{1.2cm}*{7}{>{\centering\arraybackslash}X}|m{1.2cm}*{7}{>{\centering\arraybackslash}X}}
    \hline
         \textbf{Model} &  \multicolumn{7}{c|}{\textbf{Client Test Set Accuracy}} & \textbf{Model} &  \multicolumn{7}{c}{\textbf{Client Test Set Accuracy}} \\
         \hline
         
         \hline
         
         & client1 & client2 & client3 & client4  & client5 &  client6 & avg & & client1 &   client2 & client3 & client4  & client5 &  client6 & avg\\
         \hline

            \multicolumn{8}{c|}{\textbf{DomainNet}} & \multicolumn{8}{c}{\textbf{OpenImage}}\\
         \hline
         Local & 22.22 & 8.54 & 7.67 & 28.95 & 19.16  & 16.10  & 17.64 & Local & 37.72 & 39.95 & 49.01 & 47.41 & 49.20& 41.13 & 43.97 \\
         FedAvg & 35.27 & 11.99 & 5.68& 36.99 & 22.97 & 22.33 &21.88 & FedAvg & 51.84& \textbf{52.63}& 62.70& 58.53 & 63.08 & 54.86 & 57.14 \\
         FedProx & 42.10& 11.73 & 6.29 & 42.61 & 27.53 & 25.60 & 25.33 & FedProx & 54.08& 51.30 &\textbf{63.96} & 60.53 & 63.11 & 51.19 & 57.20 \\
         FedDyn & 37.62 & 13.92 & 6.71 & 40.21 & 26.09 & 23.87 & 23.24 & FedDyn & 51.60 & 49.08& 62.75& 56.07& 59.55& 53.06& 55.22 \\
         FedCADO & 57.31 & 17.51 & 9.43  & 44.25 & \textbf{38.74}  & 38.44  & 34.28 & FedCADO & 51.66 & 48.99 & 62.41 & 55.59 & 58.86 &  52.80 & 55.05 
           \\
         FedDISC & 56.19 & 14.84 & 8.35  & 43.89 &  38.38 & 36.82 & 33.07 & FedDISC & 49.65 & 47.42 & 54.73 & 53.41 & 60.74 & 52.81 & 53.12  \\
         \rowcolor{lightgray}
         OSCAR & \textbf{66.95} &\textbf{23.25} & \textbf{10.02} & \textbf{44.54} & 34.14 & \textbf{38.97} & \textbf{37.60}& OSCAR &\textbf{55.42} & 51.14 & 63.42 & \textbf{61.12} & \textbf{68.55} & \textbf{58.11} & \textbf{59.49} \\


        \hline
                    \multicolumn{8}{c|}{\textbf{NICO++ Common}} & \multicolumn{8}{c}{\textbf{NICO++ Unique}}\\
         \hline
         Local & 54.10 & 53.95 & 42.49 & 56.68 & 53.86 & 46.14 & 51.29 & Local & 49.19 & 54.77  & 56.48 & 50.62 & 56.06 & 56.13 & 53.89 \\
         FedAvg & 58.57 & 55.36 & 44.60 & 58.63 & 55.90 & 50.27 & 54.17 & FedAvg & 69.16 &71.34 &74.22 &67.58 &\textbf{79.59} &77.14 & 73.15 \\
         FedProx & 58.63 & 52.12 & 44.96 &58.12 &54.68 & 50.43 & 53.66 & FedProx & 69.48&71.75 & 74.43 & 67.68 & 78.73 & 76.61 & 73.09 \\
         FedDyn & 62.13 & 56.62 & 48.08 & 61.76 & 57.61 & 51.60 & 56.67 & FedDyn & 66.60& 72.46& 74.84& 66.84  &77.66& 78.62 & 72.83  \\
         FedCADO & 49.21 & 58.13 & 54.63 & 54.75 & 54.64 & 47.03 & 53.06 & FedCADO & 75.13 & 70.31 & 73.60 & 68.88 &73.30 & 72.51 & 72.28 \\
         FedDISC & 51.43 & \textbf{59.45} & \textbf{56.17} & 56.82 & 52.32 & 45.64 & 53.64 &FedDISC & 74.32 & 71.25 & \textbf{75.28} & 66.79 &73.47 & 70.06  & 71.86 \\

         \rowcolor{lightgray}
          OSCAR & \textbf{59.11} & 59.32 & 52.96  & \textbf{64.04} & \textbf{62.18} &\textbf{51.70} & \textbf{58.19} & OSCAR & \textbf{75.95} & \textbf{71.32} & 75.13 & \textbf{70.14} & 75.00 & \textbf{73.93} & \textbf{73.62}\\

        \hline
    \end{tabularx}
    \label{tab:results}
\end{table*}


         
         

\paragraph{Classifier Networks} To facilitate direct comparison with existing baseline approaches, the classifier network in the main results reported is a ResNet-18. However, the synthesized data appears to have potential that more advanced backbones can utilize. The results for NICO++ Unique and NICO++ Common datasets with different classifier networks are reported in \textit{Table} \ref{tab:backbone}. The results indicate that the generated data can potentially improve the global model's optimality with an improved model architecture and may even improve more as the number of images per category increases. In the ResNet family, ResNet-101 performs the best, while the base version of the vision transformer (ViT B-16) has the best overall performance.

\begin{table}[!htb]
    \centering
    \caption{Accuracy (in \%) on the test set for OSCAR with different backbone networks at the server.}
    \begin{tabularx}{0.5\textwidth}{m{1.7cm}*{7}{>{\centering\arraybackslash}X}}
    \hline
         \textbf{Model} &  \multicolumn{7}{c}{\textbf{Client Test Set Accuracy}} \\
         \hline
         
         \hline
         
         & client1 &   client2 & client3 & client4  & client5 &  client6 & avg\\
         \hline
            \multicolumn{8}{c}{\textbf{NICO++ Unique}}\\
         \hline
         ResNet-18 & 65.42 & 71.14 & 74.02 & 68.52 & 71.00 & 70.69 & 70.15\\
         VGG-16 & 75.53 & 69.14 & 71.96  & 67.13 & 74.71 & 73.51 & 72.06\\
         ResNet-50 & 80.72 & 73.50 & 76.40  & 77.34 & 79.61 & 78.52 & 77.73\\
         ResNet-101 & 80.61 & 75.03 & 79.05  & 76.48 & 80.49 & 81.86 & 78.97\\
         DenseNet-121 & 80.51 & 77.10 & 76.93  & 75.94 & 79.61 & 78.73 & 78.17\\
         VIT B-16 & 84.53 & 77.86 & 83.49  & 79.91 & 81.08 & 82.48 & 81.58\\

         \hline

         \multicolumn{8}{c}{\textbf{NICO++ Common}}\\
         \hline

         ResNet-18 & 57.46 & 58.55 & 60.12 & 61.83 & 60.05 & 50.85 & 56.43\\
         VGG-16 & 60.31 & 62.45 & 51.68  & 63.11 & 62.04 & 54.39 & 58.95\\
         ResNet-50 & 62.19 & 65.70 & 54.44  & 67.07 & 67.78 & 55.19 & 61.76\\
         ResNet-101 & 64.04 & 69.82 & 58.01  & 69.01 & 69.32 & 57.46 & 64.16\\
         DenseNet-121 & 60.28 & 65.86 & 55.00  & 64.93 & 66.61 & 55.53 & 60.93\\
         VIT B-16 & 65.19 & 70.97 & 58.17  & 71.86 & 70.85 & 58.60 & 65.60\\
         
         \hline

        \hline
    \end{tabularx}
    \label{tab:backbone}
\end{table}

\begin{table}[!h]
    \centering
    \caption{Impact of sample count per category on OSCAR.}
    \begin{tabularx}{0.5\textwidth}{m{1cm}*{7}{>{\centering\arraybackslash}X}}
    \hline
         \textbf{Samples} &  \multicolumn{7}{c}{\textbf{Client Test Set Accuracy}} \\
         \hline
         
         \hline
         
         & client1 &   client2 & client3 & client4  & client5 &  client6 & avg\\
         \hline
            \multicolumn{8}{c}{\textbf{NICO++ Unique}}\\
         \hline
         10 & 65.42 & 71.14 & 74.02 & 68.52 & 71.00 & 70.69 & 70.15\\
         20 & 67.62 & 71.14& 74.95 & 68.94 & 74.11 & 70.90 &71.19 \\
         30 & 75.95 & 71.32 & 75.13 & 70.14 & 75.00 & 73.93 & 73.62\\
         40 & 69.26 & 72.56 &75.26 & 70.72 & 73.47 & 72.70 & 72.34 \\
         50 & 68.52 &71.44 &74.95  & 70.72 & 73.36 & 70.69 & 71.62\\

         \hline

         \multicolumn{8}{c}{\textbf{NICO++ Common}}\\
         \hline

         10 & 57.46 & 58.55 & 60.12 & 61.83 & 60.05 & 50.85 & 56.43\\
         20 & 58.03 & 57.78 & 49.52 & 62.42  & 62.13 & 52.17 & 57.06 \\
         30 & 59.11 & 59.32 & 52.96  & 64.04 & 62.18 &51.70 & 58.19\\
         40 & 58.57 & 59.10 & 61.68 & 61.90 & 61.14 & 52.17 & 57.36\\
         50 & 59.41 & 57.07& 52.08 & 62.26 & 61.42 & 53.38 & 57.93\\
         
         \hline

        \hline
    \end{tabularx}
    \label{tab:sample-count}
\end{table}

\paragraph{Number of Generated Images}
This section examines the impact of the generated dataset size on the global model performance in OSCAR. Traditionally, the increase in dataset size impacts performance positively. In this case, while the initial increase in synthesized dataset size boosts the model performance, the performance remains constant, or in some cases decreases, after a certain threshold. This may indicate that data synthesized by diffusion models may best be utilized as auxiliary data for training OSFL approaches rather than as a replacement. \textit{Table} \ref{tab:sample-count} shows the result for OSCAR with varying number of samples synthesized per category of each client.

\begin{table}[!h]
    \centering
    \caption{Total number of parameters (in millions) uploaded by each client.}
    \begin{tabularx}{0.47\textwidth}{m{1.2cm}*{5}{>{\centering\arraybackslash}X}}
    
    \hline
    Model & Local & FedAvg & FedCADO & FedDISC & OSCAR \\
    \hline
    Parameters & - & 234 & 11.69 & 4.23 &  0.03 \\
    \hline
    \end{tabularx}
    \label{tab:parameters}
\end{table}

\paragraph{Communication Analysis} As OSCAR only uploads the data encodings, it uploads the least parameters from each client. In our experiments, OSCAR uploads less than 1\% of the number of parameters compared to the evaluated SOTA models. OSCAR achieves this by eliminating the classifier training, and hence classifier uploading, and each client only communicates 512 parameters for each category. \textit{Table} \ref{tab:parameters} shows the number of parameters each client uploads. FedCADO trains a classifier model; hence, each client uploads a model with 11.69 million parameters. While FedDISC reduces the communication size by more than 60\%  compared to FedCADO at each client, the upload size from each client is still more than 100 times higher than OSCAR. 

\section{Conclusion}
In this work, we propose OSCAR, a novel one-shot federated learning approach that utilizes pre-trained vision language models and classifier-free diffusion models (DMs) to train a global model in a single communication round in FL settings. OSCAR eliminates the need for training a classifier model at each client by replacing the classifier-guided DM with a classifier-free DM in the image synthesis phase. OSCAR generates category-specific data representations for each client through BLIP and CLIP foundation models, which are communicated to the server. The server generates \textit{new} data samples and trains a global model on the generated \textit{data}. In our experiments, OSCAR reduces the communication load by reducing the client upload size more than 100X compared to state-of-the-art DM-assisted OSFL approaches while exhibiting superior performance on four benchmarking datasets. 

In future work, we want to extend OSCAR to fuse the knowledge of auxiliary generated datasets with existing learned knowledge at each client. 



\balance
\bibliography{main}

\end{document}